\documentclass[sigconf]{acmart}

\usepackage{graphicx}
\usepackage{amsmath}

\usepackage{amssymb}
\usepackage{multirow}
\usepackage{bm}
\usepackage{balance}

\usepackage{amsmath,amsfonts,bm}

\def\eqref#1{equation~\ref{#1}}

\def\1{\bm{1}}

\DeclareMathAlphabet{\mathsfit}{\encodingdefault}{\sfdefault}{m}{sl}
\SetMathAlphabet{\mathsfit}{bold}{\encodingdefault}{\sfdefault}{bx}{n}

\def\gE{{\mathcal{E}}}

\def\gL{{\mathcal{L}}}

\def\sR{{\mathbb{R}}}

\AtBeginDocument{%
  }

\copyrightyear{2023}
\acmYear{2023}
\setcopyright{acmlicensed}
\acmConference[MM '23] {Proceedings of the 31st ACM International Conference on Multimedia}{October 29--November 3, 2023}{Ottawa, ON, Canada.}
\acmBooktitle{Proceedings of the 31st ACM International Conference on Multimedia (MM '23), October 29--November 3, 2023, Ottawa, ON, Canada}
\acmPrice{15.00}
\acmISBN{979-8-4007-0108-5/23/10}
\acmDOI{10.1145/3581783.3611902}

\begin{document}
\title{Dense Object Grounding in 3D Scenes}

\author{Wencan Huang}
\authornote{Equal Contribution.}
\email{huangwencan@stu.pku.edu.cn}
\affiliation{%
  \institution{Wangxuan Institute of Computer Technology, Peking University}
  \city{Beijing}
  \country{China}
}

\author{Daizong Liu}
\authornotemark[1]
\email{dzliu@stu.pku.edu.cn}
\affiliation{%
  \institution{Wangxuan Institute of Computer Technology, Peking University}
  \city{Beijing}
  \country{China}
}

\author{Wei Hu}
\authornote{This work was supported by National Natural Science Foundation of China (61972009). Corresponding author: Wei Hu (forhuwei@pku.edu.cn).}
\email{forhuwei@pku.edu.cn}
\affiliation{%
  \institution{Wangxuan Institute of Computer Technology, Peking University}
  \city{Beijing}
  \country{China}
}

\renewcommand{\shortauthors}{Wencan Huang, Daizong Liu \& Wei Hu}

\begin{abstract}
    Localizing objects in 3D scenes according to the semantics of a given natural language is a fundamental yet important task in the field of multimedia understanding, which benefits various real-world applications such as robotics and autonomous driving. 
    However, the majority of existing 3D object grounding methods are restricted to a single-sentence input describing an individual object, which cannot comprehend and reason more contextualized descriptions of multiple objects in more practical 3D cases. 
    To this end, we introduce a new challenging task, called 3D Dense Object Grounding (3D DOG), to jointly localize multiple objects described in a more complicated paragraph rather than a single sentence.
    Instead of naively localizing each sentence-guided object independently, we found that dense objects described in the same paragraph are often semantically related and spatially located in a focused region of the 3D scene. To explore such semantic and spatial relationships of densely referred objects for more accurate localization, we propose a novel Stacked Transformer based framework for 3D DOG, named 3DOGSFormer. Specifically, we first devise a contextual query-driven local transformer decoder to generate initial grounding proposals for each target object. The design of these contextual queries enables the model to capture linguistic semantic relationships of objects in the paragraph in a lightweight manner. Then, we employ a proposal-guided global transformer decoder that exploits the local object features to learn their correlation for further refining initial grounding proposals. In particular, we develop two types of proposal-guided attention layers to encode both explicit and implicit pairwise spatial relations to enhance 3D relation understanding. Extensive experiments on three challenging benchmarks (Nr3D, Sr3D, and ScanRefer) show that our proposed 3DOGSFormer outperforms state-of-the-art 3D single-object grounding methods and their dense-object variants by significant margins.
\end{abstract}

  

\maketitle


\begin{figure}[!tbp]
    \begin{center}
    \includegraphics[width=0.475\textwidth]{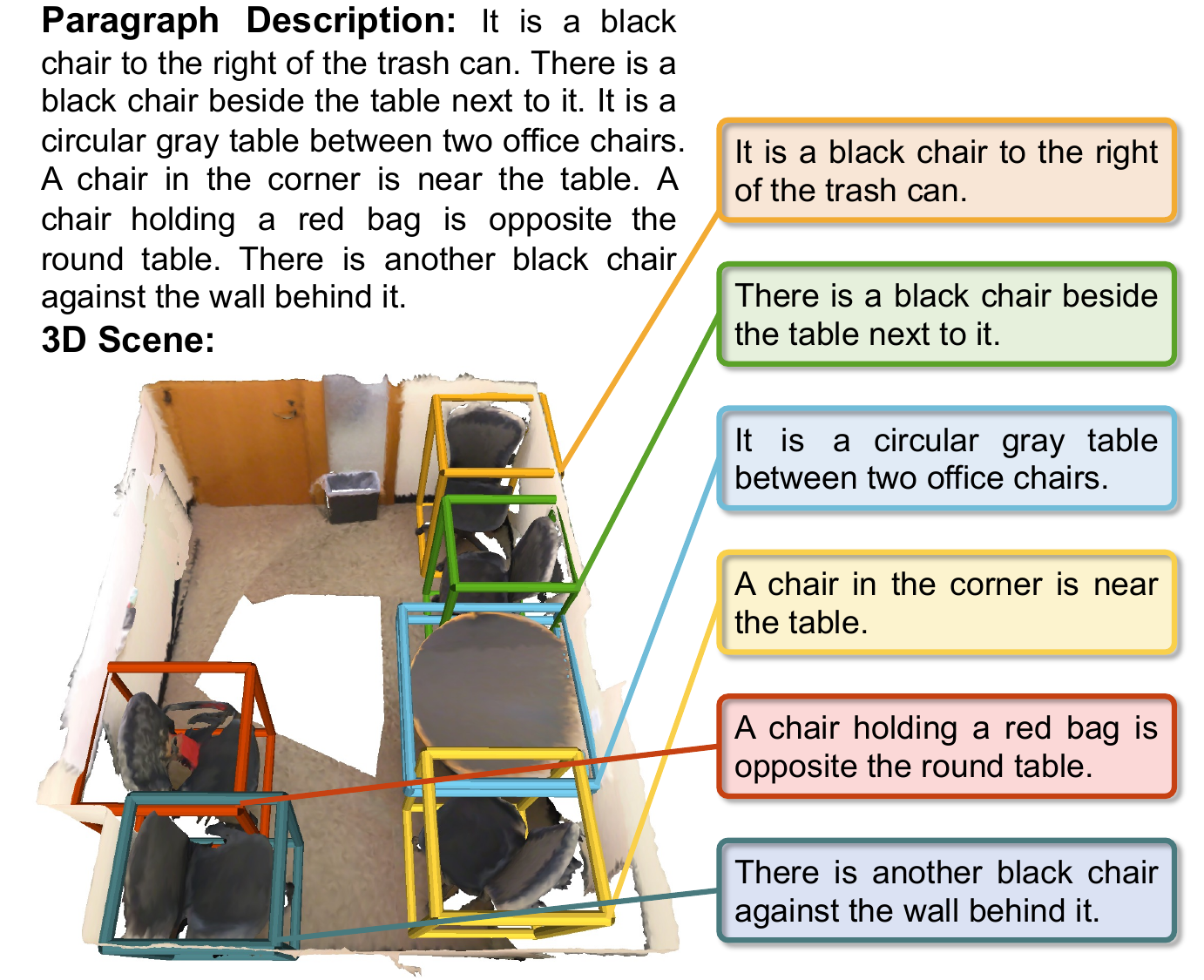}
    \end{center}
    \caption{Dense Object Grounding in 3D Scenes. Given a paragraph description, 3D dense object grounding (3D DOG) aims to jointly localize described multiple objects in a 3D scene. }
    \label{fig:intro}
\end{figure}
\section{Introduction}
Grounding natural language in visual contents is a fundamental yet essential task in the multimedia understanding field. Traditional object grounding in 2D images aims to localize the object described by the given referring expression in an image, which has attracted much attention and made great progress \cite{yu2016modeling,mao2016generation,liu2019clevr,chen2020cops,chen2019touchdown,qi2020reverie,qiao2020referring,liu2021context,liu2020jointly}. Recently, researchers begin to explore real-world 3D object grounding \cite{chen2020scanrefer,achlioptas2020referit3d}, where the target object referred by a sentence should be identified in a more complicated 3D scene. Compared to 2D object grounding, 3D object grounding is more challenging since it requires disambiguating more variant and complex spatial relationships and localizing the object referred by the text in 3D scenes with several same-class distractors.

The paradigm of previous studies on 3D object grounding is to simply localize each individual object referred by a single free-form sentence in a 3D scene. Existing 3D grounding approaches can be mainly categorized into two groups: (1) \textit{Top-down} methods: these works directly follow a detection-then-matching two-stage framework \cite{feng2021free,he2021transrefer3d,yang2021sat,yuan2021instancerefer,zhao20213dvg,huang2022multi,bakr2022look,chen2022language,roh2022languagerefer}, which first employ pre-trained 3D object detectors to generate multiple object proposals, then select the best proposal according to cross-modal similarity scores with the given sentence. 
However, these methods severely rely on the quality of the detected proposals and are often very time-consuming.
(2) \textit{Bottom-up} methods: instead of using complex proposals, these methods simply incorporate the sentence features and point-level visual representations in an early-fusion manner to directly regress the bounding box at a single stage \cite{liu2021refer,luo20223d}. 
Although the above two types of methods have achieved significant progress in recent years, they are limited by the single-sentence input and are not suitable for comprehending and reasoning more contextualized descriptions of multiple objects in complicated 3D scenes.
\textit{However, such a paragraph of multiple sentences that describe several objects in a specific region of a 3D scene is natural and practical in real-world applications such as robotics and autonomous driving.}
For example, as shown in Figure \ref{fig:intro}, humans may be interested in multiple objects located in a focused region of the scene.
Although multiple objects with the same category like ``\textit{chair}'' may appear in the scene, humans may only be interested in one of them. 
Therefore, they can utilize a paragraph consisting of multiple sentences to describe not only the most concerned object but also its context-related objects, to better avoid the ambiguity. 

To localize dense objects referred by a paragraph in a 3D scene, a straightforward idea is to apply well-studied 3D single-object grounding models \cite{huang2022multi,bakr2022look,chen2022language,roh2022languagerefer,luo20223d} to each individual sentence in the paragraph and integrate their results for dense grounding. 
However, directly applying them to the multiple-sentence setting may suffer from two critical issues: (1) \textit{Firstly}, this solution is naive and only considers a single sentence, failing to learn the contextual semantic relationships among multiple sentences. These contextual semantic relations are vital clues for accurate paragraph comprehension and precise localization of dense objects. For instance, as shown in Figure \ref{fig:intro}, grounding sentence ``\textit{There is another black chair against the wall behind it}'' requires understanding the coreference relationship between the anaphor ``\textit{it}'' and the target object ``\textit{chair}'' referred by another sentence ``\textit{A chair holding a red bag is opposite the round table}''. 
(2) \textit{Secondly}, this solution fails to leverage the spatial correlations between locations of dense objects described in the same paragraph for cross-modal spatial reasoning. 
The existence of such spatial correlations is due to the fact that humans are likely to refer linguistically to multiple objects located in a focused region of a 3D scene rather than describe randomly placed ones.
Ignoring such spatial relations of dense target objects may lead to inferior performance in precisely finding the location of each one. 

Motivated by the above observations, in this paper, we mainly focus on addressing the challenging 3D Dense Object Grounding (3D DOG) task. Given a 3D point cloud scene and a paragraph of sentence descriptions, the goal of 3D DOG is to jointly localize dense objects described by these sentences. Rather than localizing each object independently, we explore the {\it semantic and spatial relationships} of densely referred objects for accurate localization. To achieve this, we propose a novel two-phase framework for 3D DOG based on Stacked Transformers (named 3DOGSFormer) due to the powerful relation modeling capability of the well-known transformer architecture \cite{vaswani2017attention}. Specifically, the proposed 3DOGSFormer consists of one standard transformer encoder for 3D scene encoding and two types of transformer decoders for the two-phase grounding pipeline. In the first phase, we devise a shared contextual query-driven local transformer decoder to generate initial grounding proposals for every single sentence in the paragraph description. We propose a scene-aware context aggregation and propagation (SCAP) module to generate contextual queries for each sentence, which enables to capture the semantic relations between multiple sentences in a lightweight fashion. 
In the second phase, we further develop a novel proposal-guided global transformer decoder to gather information from local proposal features and learn to refine the initial grounding proposals via cross-modal spatial reasoning. To be specific, we design a stack of interlaced proposal-guided self-attention (PGSA) and cross-attention (PGCA) layers to encode both explicit and implicit pairwise spatial relations for all object-object and object-point pairs, so as to enhance 3D spatial relation understanding of densely referred objects. 

Our main contributions can be summarized as follows:
\begin{itemize}
    \item We propose a new 3D DOG task to explore practical yet challenging 3D dense object grounding based on complicated paragraph descriptions.
    \item We develop a novel 3DOGSFormer to tackle this 3D DOG task in a two-phase grounding pipeline, where we design a contextual query-driven local transformer decoder to capture the semantic relationships within the paragraph efficiently, as well as a proposal-guided global transformer decoder to enhance the 3D spatial relation understanding.
    \item Extensive experiments on three challenging benchmarks (Nr3D \cite{achlioptas2020referit3d}, Sr3D \cite{achlioptas2020referit3d}, and ScanRefer \cite{chen2020scanrefer}) show that the proposed 3DOGSFormer outperforms state-of-the-art 3D single-object grounding methods and their dense-object variants by significant margins.
\end{itemize}


\section{Related Work}
\subsection{3D Single Object Grounding}
The 3D object grounding task \cite{chen2020scanrefer,achlioptas2020referit3d} aims to localize objects in 3D point clouds given a sentence. Existing approaches can be mainly categorized into two groups, namely top-down and bottom-up frameworks. The top-down methods \cite{feng2021free,he2021transrefer3d,yang2021sat,yuan2021instancerefer,zhao20213dvg,huang2022multi,bakr2022look,chen2022language,roh2022languagerefer} follow a detection-then-matching two-stage pipeline, which first obtain the features of the query sentence and candidate point cloud objects independently by a pre-trained language model \cite{kenton2019bert} and a pre-trained 3D detector \cite{liu2021group,qi2019deep} or segmentor \cite{chen2021hierarchical,jiang2020pointgroup,vu2022softgroup}, then employ various cross-modal fusion or matching mechanisms to select the best-matched object according to the sentence. 
Graph-based approaches \cite{achlioptas2020referit3d,huang2021text,yuan2021instancerefer,feng2021free} and Transformer-based attention mechanisms \cite{he2021transrefer3d,zhao20213dvg,yang2021sat,huang2022multi,bakr2022look,chen2022language,roh2022languagerefer} are widely adopted for the multi-module feature fusion in the matching stage.
The obvious drawback of the top-down methods is that they severely rely on the quality of the detected proposals and are very time-consuming.
By contrast, bottom-up models \cite{liu2021refer,luo20223d} incorporate the sentence features and voxel- or point-level visual representations in an early-fusion manner to directly regress the bounding box at a single stage, which are more flexible to identify various text-concerned objects. Typically, 3D-SPS \cite{luo20223d} employs textual features to guide visual keypoint selection and progressively localizes objects. Recently, BEAUTY-DETR \cite{jain2022bottom} develops a Transformer-based bottom-up top-down architecture, which combines the advantages of the above two pipelines. Although existing methods have made great attempts to solve the 3D object grounding problem, they are restricted to the single-sentence input describing an individual object. In this paper, we propose a new but challenging 3D DOG task to explore 3D dense object grounding based on complicated paragraph descriptions.

\subsection{3D Dense Object Understanding}
Understanding dense objects in 3D scenes has raised great interest among researchers in recent years. Scan2Cap \cite{chen2021scan2cap} introduces the task of 3D dense captioning, which aims to jointly localize and describe dense objects in a 3D scene. To tackle this task, it builds a message-passing graph network to mine the relations among objects in a 3D scene. MORE \cite{jiao2022more} further takes multi-order relations into account to learn richer 3D object relation features. SpaCap3D \cite{DBLP:conf/ijcai/0007ZY022} uses a spatiality-guided transformer to learn the contribution of surrounding objects to the target object for 3D dense captioning. 3DJCG \cite{cai20223djcg} and UniT3D \cite{chen2022unit3d} develop unified transformer-based frameworks for both 3D dense captioning and 3D object grounding. X-Trans2Cap \cite{yuan2022x} introduces additional 2D prior information to improve 3D dense captioning with knowledge transfer. \cite{zhong2022contextual} shifts attention to contextual information for the perception of non-object information. Recently, Vote2Cap-DETR \cite{chen2023end} proposes a one-stage model for 3D dense captioning that detects and generates captions in parallel. The 3D dense captioning task in these papers is to describe dense objects in the 3D scene with a paragraph of multiple sentences. In contrast, our 3D dense object grounding task can be viewed as the inverse problem of 3D dense captioning. 

The methods most similar to our work are EDA \cite{wu2022eda}, PhraseRefer \cite{yuan2022toward}, and ScanEnts3D \cite{z_abdelreheem2022scanents3d}, which propose to ground not only the target object but also all auxiliary objects mentioned in the referential utterance to improve 3D object grounding. However, these methods are still limited in the single-sentence setting and cannot deal with the challenging 3D dense object grounding task, which requires to comprehend the semantic and spatial relationships of dense objects described by multiple sentences in the same paragraph.

\subsection{Transformers in 2D and 3D Scenes}
Transformer \cite{vaswani2017attention} has achieved marvelous success in most 2D computer vision tasks \cite{DBLP:conf/iclr/DosovitskiyB0WZ21,liu2021swin}, such as visual grounding \cite{kamath2021mdetr,yang2022tubedetr}, image captioning \cite{cornia2020meshed}, and object detection \cite{carion2020end,DBLP:conf/iclr/ZhuSLLWD21}. In particular
, DETR \cite{carion2020end} introduces a new query-based \cite{sun2021sparse,DBLP:conf/iclr/ZhuSLLWD21} paradigm for object detection, which employs a set of object queries as candidates and feeds them to the Transformer decoder for parallel detection. Beyond 2D field, the DETR architecture has been extended for various 3D vision tasks such as 3D object detection \cite{misra2021end,liu2021group,zhu2022conquer}, 3D instance segmentation \cite{liu20223d,sun2022superpoint}, 3D object grounding \cite{jain2022bottom}, and 3D dense captioning \cite{chen2023end}. In our work, we extend the DETR architecture to build a new 3DOGSFormer framework for 3D dense object grounding, which employs stacked transformer decoders to localize dense objects in parallel. Additionally, we leverage contextual queries for efficient semantic relation modeling and develop proposal-guided attention layers to enhance the 3D spatial relation understanding.


\section{Method}
\begin{figure*}[!t]
    \begin{center}
    \includegraphics[width=0.9\textwidth]{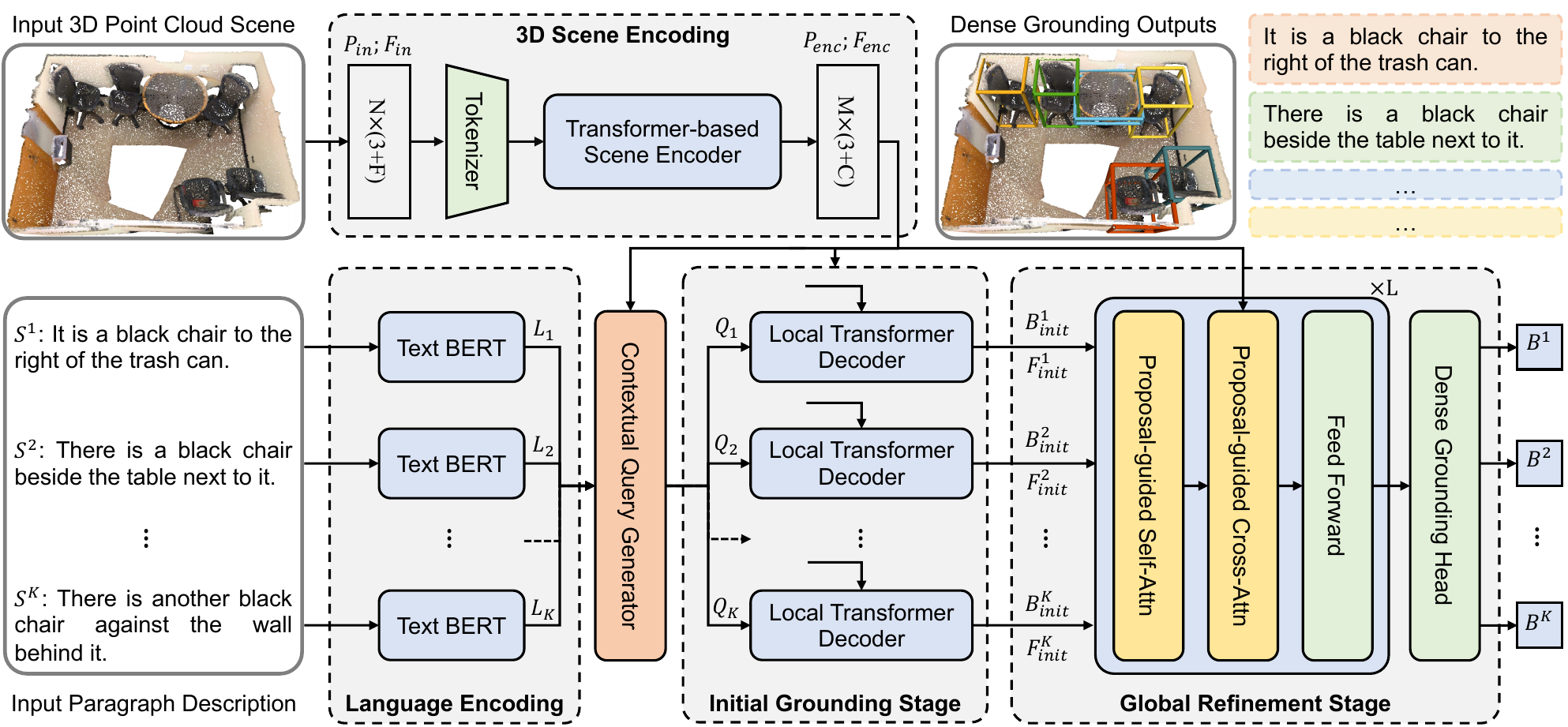}
    \end{center}
    \caption{The pipeline of our 3DOGSFormer. After the 3D scene and paragraph encoding, we initialize the grounding proposals with a local transformer decoder, then refine them with a global transformer decoder.
    With the help of contextual queries and proposal-guided attention layers, the semantic and spatial relations of densely referred objects can be efficiently captured.
    }
    \label{fig:model}
\end{figure*}

Given a 3D scene and a paragraph description of multiple sentences, the goal of 3D DOG is to jointly localize dense objects described by these sentences. Generally, we represent the input 3D scene as a point cloud $PC=\left[P_{in}; F_{in}\right] \in \sR^{N\times (3+F)}$, where $P_{in} \in \sR^{N\times 3}$ is the absolute locations for each point, and $F_{in}\in \sR^{N\times F}$ is additional input feature for each point, such as \textit{color}, \textit{normal}, \textit{height}, or \textit{multiview feature} introduced by \cite{chen2020scanrefer}. We denote the input paragraph description containing $K$ sentences as $S$, and the $k$-th sentence description $S^k$ is presented as $S^k=\{s_{k,i}\}^{T_k}_{i=1}$, where $s_{k,i}$ represents the $i$-th word and $T_k$ denotes the total number of words. The expected output for the $k$-th sentence is the bounding box $B^k$, representing an estimated location of the target object corresponding to the semantics of this sentence in the 3D scene.

To tackle the challenging 3D DOG task, we propose a novel Stacked Transformer based framework called 3DOGSFormer to exploit both semantic and spatial relationships among objects in the scene. As shown in Figure \ref{fig:model}, we first adopt a 3DETR \cite{misra2021end} encoder as our scene encoder, and a BERT model \cite{kenton2019bert} as the paragraph encoder. Then we develop a shared contextual query-driven local transformer decoder to generate initial grounding proposals for each input sentence. At last, we further devise a proposal-guided global transformer decoder that exploits the obtained local object features to learn their correlation to refine their bounding box (bbox) proposals for final grounding.

\subsection{3D Scene and Paragraph Encoder}

\noindent \textbf{3D Scene Encoder.} We exploit the powerful 3DETR \cite{misra2021end} encoder to extract 3D visual features from the input 3D point cloud scene. During the 3D feature encoding, the input $PC$ is first tokenized with a set-abstraction layer \cite{qi2017pointnet++}. Then, point tokens are fed into a masked transformer encoder with a set-abstraction layer followed by another two encoder layers. We denote the encoded scene tokens as $\left[P_{enc}; F_{enc}\right] \in \sR^{M\times (3+C)}$.

\noindent \textbf{Paragraph Encoder.} Following \cite{chen2022language,bakr2022look,roh2022languagerefer}, we adopt the BERT model \cite{kenton2019bert} as the paragraph encoder. Specifically, the BERT model is shared across all input sentences in the paragraph description and extracts feature vectors for each sentence separately. Given the $k$-th sentence $S^k$ with $T_k$ words, we embed them into $D$-dimensional feature vectors $L_k=\left(l^k_{s}, l^k_{1}, \cdots, l^k_{T_k}\right)$, where $l^k_{s}$ is the sentence-level feature, $l^k_{i}$ is the feature of the $i$-th word.

\subsection{Contextual Query Generator}

\begin{figure}[!tb]
    \begin{center}
    \includegraphics[width=0.35\textwidth]{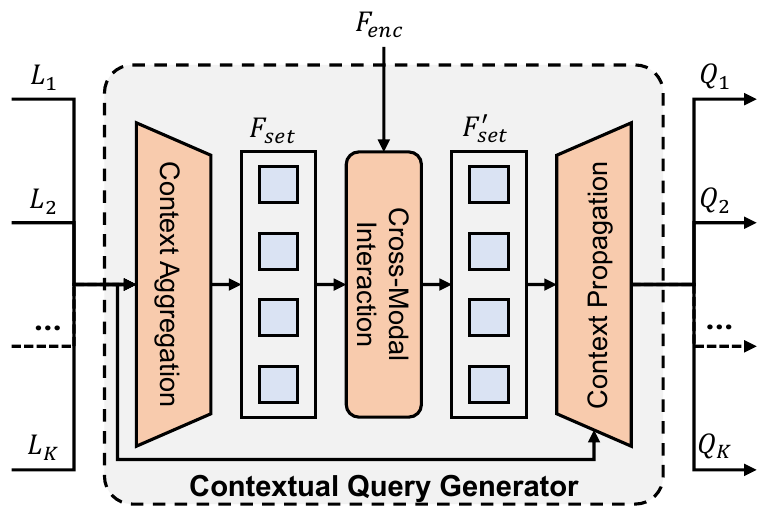}
    \end{center}
    \caption{
    The contextual query generator, which includes context aggregation, cross-modal interaction, and context propagation modules.
    }
    \label{fig:module1}
\end{figure}

We devise a contextual query generator to generate query embeddings for each input sentence. These queries should enable the local transformer decoder to generate the desired initial grounding proposals. To achieve this, the queries for each sentence must learn the following crucial information: (1) the complete semantics of the corresponding sentence; (2) contextual semantics among multiple sentences in the paragraph, which are vital clues for accurately understanding each individual sentence; (3) the knowledge of the 3D spatial location and the scene-related visual information, which are proved \cite{misra2021end,chen2023end} to be effective in multi-modal semantic alignment and reasoning. To comply with the requirement (1), we directly utilize the encoded language feature vector $L_k$ as the initial query embeddings for $S^k$. Then we design a novel scene-aware context aggregation and propagation (SCAP) mechanism to update the initial queries so as to meet the requirement (2) and (3). As shown in Figure \ref{fig:module1}, the SCAP module includes context aggregation, cross-modal interaction, and context propagation modules.

\noindent\textbf{Context Aggregation.}
The high computational complexity is the biggest challenge when encoding paragraph-level contextual information into the queries for each sentence. To address this issue, we devise a context aggregation module to adaptively aggregate the contextual information in the long paragraph into a compact set of features with a much fewer number of tokens, which enables efficient contextual learning. Specifically, we first concatenate the encoded language feature vectors $\left[L_1, L_2, \cdots, L_K\right]$ to produce a paragraph feature vector $L \in \sR^{T\times D}$, where $T$ is the number of tokens in the paragraph. We then employ $N_s$ learnable queries $Q_{set}\in \sR^{N_s\times C}$ to perceive the contextual information in $L$ and aggregate it into a compact set $F_{set} \in \sR^{N_s\times C}$ via a plain multi-head cross-attention layer \cite{vaswani2017attention}, as $L^{\prime} = LW_{l} + e_p, F_{set} = \textup{CrossAttn}(Q_{set}, L^{\prime}, L^{\prime})$, where $W_l\in \sR^{D\times C}$ is the projection matrix and $e_p$ is the learnable positional embedding vector.

\noindent\textbf{Cross-Modal Interaction.}
To inject spatial knowledge and 3D scene-related information into the contextual query features, we then use a Co-Attention module \cite{lu2016hierarchical} to fuse the compact set of semantic features and the encoded 3D scene features. Specifically, given the input feature vectors $F_{set}$ and $F_{enc}$, we conduct multi-stage co-attention in both vision-to-language and language-to-vision directions and generate a compact set of fused features $F_{set}^{\prime} \in \sR^{N_s\times C}$ based on the language-side attention results from all stages, as: 
\begin{equation}
F_{set}^{i+1}, F_{enc}^{i+1} = \textup{CoAttn}(F_{set}^{i}, F_{enc}^{i}), i \in \{0,1,2\},
\end{equation}
\begin{equation}
F_{set}^{\prime} = \left[F_{set}^{0};F_{set}^{1};F_{set}^{2};F_{set}^{3}\right] W_f,
\end{equation}
where $F_{set}^{i}$ and $F_{enc}^{i}$ are hidden features from the $i$-th co-attention layer, $F_{set}^{0}=F_{set}$, $F_{enc}^{0}=F_{enc}$, $[;]$ denotes concatenation and $W_f\in \sR^{4C \times C}$ is the projection matrix. The cross-modality interaction is achieved by the $\textup{CoAttn}$ module, which computes the hidden vectors of one modality by attending to the other modality, given by: 
\begin{equation}
A_{i} = (F_{set}^{i} W^{i}_{set})(F_{enc}^{i} W^{i}_{enc})^T / \sqrt{d},
\end{equation}
\begin{equation}
F_{set}^{i+1} = \textup{softmax}(A_i)F_{enc}^{i}\hat{W}^{i}_{enc}, F_{enc}^{i+1} = \textup{softmax}(A_i^T)F_{set}^{i}\hat{W}^{i}_{set},
\end{equation}
where $d$ is the dimension of the embedding space and $W^{i}_{set}, W^{i}_{enc}\in\sR^{C\times d}$, $\hat{W}^{i}_{enc}, \hat{W}^{i}_{set}\in\sR^{d\times C}$ are all projection matrices.

\noindent\textbf{Context Propagation.}
At last, we selectively propagate the aggregated multi-modal global information to each sentence for the contextual query generation. In detail, given the language feature vector $L_k$ of the $k$-th sentence, we leverage a multi-head cross-attention layer \cite{vaswani2017attention} to retrieve its contextual queries from the compact set of fused features, as $L_k^{\prime} = L_k W_q, Q_k = \textup{CrossAttn}(L_k^{\prime}, F_{set}^{\prime}, F_{set}^{\prime}) + L_k^{\prime}$, where $W_q \in \sR^{D\times C}$ is the projection matrix and $Q_k\in \sR^{(T_k+1) \times C}$ is the output contextual queries for $S^k$.

\subsection{Local Transformer Decoder}
We devise a local transformer decoder to generate initial grounding proposals for each sentence independently. The local transformer decoder is based on a standard Transformer decoder \cite{vaswani2017attention} and is shared across all sentences. For the $k$-th sentence, it takes as input the encoded 3D scene features and a sequence of contextual queries $Q_k = \left(q^k_s, q^k_1, \cdots, q^k_{T_k}\right)$ to produce a list of features that are then used to predict 3D-bounding boxes via a shared FFN-based grounding head. In our framework, the query embedding $q^k_i$ represents a potential object mentioned by the $i$-th word in the sentence, and $q^k_s$ is the sentence-level query embedding representing the target object described by $S^k$. At last, the predicted 3D-bounding box of $q^k_s$ and the corresponding feature are obtained as the initial grounding proposal $B^k_{init}$ and the contextual feature $F^k_{init}$. Following 3DETR \cite{misra2021end} and 3D-SPS \cite{luo20223d}, 3D-bounding box estimation is formulated as box center and box size estimation. We use positional embeddings in the decoder which operates on both the 3D scene features and the query embeddings. 

\subsection{Global Transformer Decoder}
To capture the 3D spatial relationships among multiple target objects and refine the initial grounding proposals, we further develop a global transformer decoder, which consists of L proposal-guided transformer layers. Each transformer layer comprises a proposal-guided self-attention layer, a proposal-guided cross-attention layer, and a feed-forward neural network (FFN). Assuming $B^k_{l} \in \sR^6$ and $F^k_{l}\in \sR^C$ are the input proposal and the contextual feature for the target object described by $S^k$ before the $(l+1)$-th transformer layer, and $B^k_{l}$ consists of the object center $c^k_{l} \in \sR^3$ and the object size $s^k_{l} \in \sR^3$, we first use a linear projection layer to obtain the absolute 3D location feature as $p^k = W_p B^k_{l} \in \sR^C$ which is added to the contextual feature $F^k_{l}$ to enhance the spatial information. Next, we utilize the proposal-guided self-attention layer to exploit the contextual 3D spatial relationships among dense referred objects and generate the enhanced proposal feature $\hat{F}^k_{l}$ for the $k$-th object. After that, our novel proposal-guided cross-attention layer takes $\left[B^k_{l}; \hat{F}^k_{l}\right]$ as queries and 3D scene features $\left[P_{enc}; F_{enc}\right]$ as keys and values to learn the spatial relations between the initial proposals and the correct locations of each target object in the 3D scene so as to generate the refined proposal feature. 

\begin{figure}[!tb]
    \begin{center}
    \includegraphics[width=0.3\textwidth]{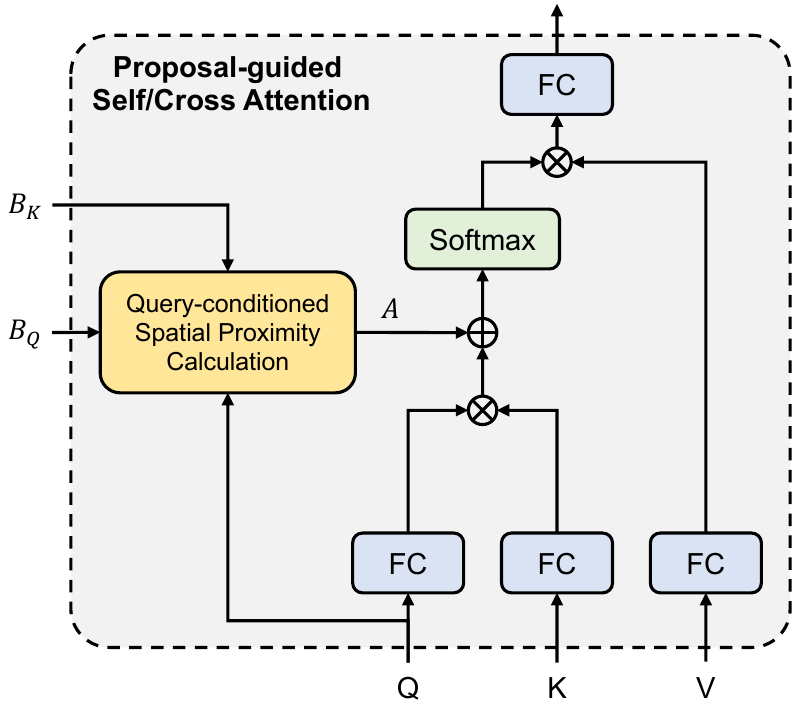}
    \end{center}
    \caption{The proposal-guided attention module, where the attention matrix is augmented by the query-conditioned spatial proximity matrix.
    }
    \label{fig:module2}
\end{figure}

Specifically, the aforementioned proposal-guided self- or cross-attention modules are extended from the standard attention mechanism \cite{vaswani2017attention}. As shown in Figure \ref{fig:module2}, we use a \textbf{Query-Conditioned Spatial Proximity Calculation (QCSPC)} module to guide the traditional attention calculation by adding a spatial proximity matrix to the original attention matrix, given the additional input proposals $B_Q=\left[c_Q;s_Q\right]\in\sR^{N_Q\times 6}$ and $B_K=\left[c_K;s_K\right]\in\sR^{N_K\times 6}$. Note that we set $B_K=\left[P_{enc}; \varepsilon\right]\in\sR^{M\times 6}$ in the proposal-guided cross-attention, where $\varepsilon$ is a hyperparameter. To better capture the object-object or object-point pairwise 3D spatial relations, our QCSPC module computes the explicit spatial proximity matrix $A^E$ and the implicit spatial matrix $A^I$, which are added together to form the spatial proximity matrix as $A=A^E+A^I$. 

\noindent\textup{\textbf{Explicit Spatial Proximity Calculation.}} We exploit a pairwise spatial feature $f^E_{ij}\in\sR^5,i\in\left[1,N_Q\right],j\in\left[1,N_K\right]$ to model the spatial relations explicitly. For each pair of proposals $(B^i_Q, B^j_K)$, we compute their Euclidean distance $d_{ij}=\Vert c^i_Q-c^j_K\Vert_2$ as well as horizontal and vertical angles $\theta_h, \theta_v$ of the line connecting their centers $c^i_Q$ and $c^j_K$. The explicit pairwise spatial feature $f^E_{ij}$ is then defined as $f^E_{ij} = \left[ d_{ij}, \sin(\theta_h), \cos(\theta_h), \sin(\theta_v), \cos(\theta_v) \right]$.
We generate a query-conditioned weight $g^E_i$ to select relevant spatial relations for each query proposal as $g^E_i=Q^iW^E$, where $Q^i\in\sR^{C}$ is the $i$-th input query and $W^E\in\sR^{C\times 5}$ is a learnable parameter. We then define the explicit spatial relevance as $A^E_{ij} = g^E_i \cdot f^E_{ij}$.

\noindent\textup{\textbf{Implicit Spatial Proximity Calculation.}} We propose to extract implicit pairwise spatial relation features to complement the explicit spatial relations. Specifically, for each pair of 3D object proposals $(B^i_Q, B^j_K)$, we first obtain raw XYZ points inside these object proposals from the 3D scene as $(O^i_Q, O^j_K)$, and then we encode them using PointNet++ \cite{qi2017pointnet++} networks $\gE_Q$ and $\gE_K$, respectively, to extract their spatial relational feature vector. Next, we concatenate the per-proposal encoding for $B^i_Q$ and $B^j_K$ and employ a 2-layer multi-layer perceptron to extract the $C$-dimensional implicit pairwise spatial relation feature for the proposal pair:
\begin{equation}
    f^I_{ij} = \textup{MLP}(\left[\gE_Q(O^i_Q);\gE_K(O^j_K)\right]).
\end{equation}
A query-conditioned weight $g^I_i=Q^iW^I$ is then generated, where $W^I\in\sR^{C\times C}$ is a learnable parameter. At last, we define the implicit spatial relevance for $(B^i_Q, B^j_K)$ as $A^I_{ij} = g^I_i \cdot f^I_{ij}$.

\noindent\textup{\textbf{Focused Region Constraint.}} Considering that humans are likely to describe multiple objects located in a focused region of a 3D scene, we propose to calculate an additional term $A^F \in \sR^{N_Q\times N_K}$ to adaptively constrain cross-attention within the currently focused region of the 3D scene. We propose to use the center point of all input query proposals as the focused point, \emph{i.e.}, $c_F = \sum_{i=1}^{N_Q} c^i_Q / N_Q$, and compute the radius of the focused region as $R_F = \max_{i=1}^{N_Q}\Vert c_F-c^i_Q\Vert_2$. The additional term $A^F$ filters points in the 3D scene by calculating their Euclidean distances to the focused point and comparing them with a threshold $\tau R_F$, as:
\begin{equation}
    A^F_{ij} = \begin{cases}
        0 & \textup{if}\quad \Vert c_F-c^j_K\Vert_2 < \tau R_F\\
        -\infty & \textup{otherwise}
    \end{cases}.
\end{equation}
Empirically, we set $\tau$ to 2. We add the term to the spatial proximity matrix when we conduct the proposal-guided cross-attention as $A=A^E+A^I+A^F$.

After the proposal-guided self- and cross-attention, the final FFN uses two fully connected layers to further encode the output tokens. Given the output embedding $F^k_L$ from the last transformer layer, we use a two-layer feed-forward neural network as the dense grounding head to predict a bounding box around every referred object. The prediction consists of a center offset $\Delta c^k_L$ and a size offset $\Delta s^k_L$, which are added to the initial proposal, \emph{i.e.}, $B^k=\left[c^k_{L-1}+\Delta c^k_L; s^k_{L-1}+\Delta s^k_L\right]$.

\subsection{Training Objectives}

\noindent\textup{\textbf{Initial Grounding Loss.}} In the initial grounding stage, we directly obtain the predicted 3D bounding boxes $\{\hat{B}_{init}\}$ for each referring sentence in the paragraph from the local transformer decoder. To supervise the training, we use a weighted sum of an L1 loss and a Generalized IOU loss \cite{rezatofighi2019generalized} following the previous work \cite{jain2022bottom}, as $\gL_{init}=\lambda_{iou}\gL_{iou}(\hat{B}_{init}, B_{init})+\lambda_{L1}||\hat{B}_{init}-B_{init}||_1$. The $\lambda_{iou}$ and $\lambda_{L1}$ control the relative weighting of the two losses in the initial grounding objective.

\noindent\textup{\textbf{Refinement Loss with Iterative Prediction.}} Considering the slow convergence of transformer-based model \cite{carion2020end}, we feed every layer output features $F^k_l$ of the global transformer decoder into the shared dense grounding head to generate proposals $B^k_l=B^k_{l-1}+\Delta B^k_l$, and feed the generated proposals into the next transformer decoder layer as its input. During training, we use a weighted sum of a L1-based center offset regression loss $\gL_{cent\_reg}$ and a L1-based size offset regression loss $\gL_{size\_reg}$ to supervise each layer's output, as $\gL_{refine}=\lambda_{cent}\gL_{cent\_reg}+\lambda_{size}\gL_{size\_reg}$.

\noindent\textup{\textbf{Pretraining the Local Transformer.}} Although modules in 3DOGSFormer can be trained end-to-end from scratch, we found that a simple pretrain-then-finetune strategy can stabilize the training process and improve the performance. Specifically, we first pretrain the encoders and the local transformer decoder without the refinement module and the contextual query generator. We then train the contextual query generator and finetune the encoders and the local decoder without the refinement module. At last, we train the global transformer decoder and finetune all other modules in 3DOGSFormer with the total loss $\gL=\lambda_{refine}\gL_{refine}+\lambda_{init}\gL_{init}$.

\noindent\textup{\textbf{Data Augmentation.}} During training, we use data augmentation strategies to alleviate the overfitting issue. Concretely, following previous work \cite{zhao20213dvg}, we randomly erase some words in each sentence of the input paragraph before the BERT encoder to alleviate the issue that the grounding model is mainly decided by the prominent parts of the sentences. Additionally, we add Gaussian noise to the initial grounding proposals before every layer of the global transformer decoder to learn more robust features for more accurate grounding.


\section{Experiments}

\subsection{Experimental Settings}
\noindent\textbf{Datasets.}
We evaluate the performance on the commonly used datasets in 3D single-object grounding, ScanRefer \cite{chen2020scanrefer} and Nr3D/Sr3D \cite{achlioptas2020referit3d}, and adopt the same evaluation metrics for ease of comparisons. \textbf{ScanRefer} \cite{chen2020scanrefer} is built on 3D scenes from ScanNet \cite{dai2017scannet}. ScanRefer has 36,665 free-form language annotations describing 7,875 objects from 562 3D scenes for training, and evaluates on 9,508 sentences for 2,068 objects from 141 3D scenes. According to whether the target object is a unique object class in the scene, the dataset is divided into a “unique” and a “multiple” subset in evaluation. The evaluation metric of the dataset is the Acc@$m$IoU, which means the fraction of descriptions whose predicted box overlaps the ground truth with IoU > $m$, where $m\in \{0.25,0.5\}$. \textbf{Nr3D/Sr3D} \cite{achlioptas2020referit3d} is also proposed based on ScanNet \cite{dai2017scannet}, with Nr3D containing 41,503 human-written sentences similar to ScanRefer’s text annotation and Sr3D including 83,572 synthetic expressions generated by templates. Sentences in Nr3D and Sr3D are split into “easy” and “hard” subsets in evaluation based on whether the target object contains more than one same-class distractor. GT boxes for all candidate objects in the scene are provided by these datasets. The metric is the accuracy of selecting the target bounding box among the proposals.

\noindent\textbf{Implementation Details.}
For the model architecture, we set the dimension $C=256$ and use 8 heads for all the transformer layers. The text encoding module is a three-layer transformer initialized from BERT \cite{kenton2019bert}, and both the transformer encoder and decoders contain 4 attention blocks. We use the point cloud tokenizer to subsample $M=1024$ points. We set the size of compact set $N_s$ to 64.
We adopt the pretrain-then-finetune training process.
In each training stage, we utilize rotation augmentation to increase the viewpoint invariance. The hyper-parameters $\lambda_{iou}$, $\lambda_{L1}$, $\lambda_{cent}$, $\lambda_{size}$, $\lambda_{refine}$, and $\lambda_{init}$ are empirically set to 1.0, 1.0, 1.0, 1.0, 1.0, and 0.05, respectively.
The hyper-parameter $\varepsilon$ in the proposal-guided cross-attention is set to 0.01. During the end-to-end training, we use the AdamW algorithm \cite{loshchilov2017decoupled} to optimize the loss function with the initial learning rate of $5\times 10^{-4}$ and the batch size of 2.

\begin{table*}[!t]
\center
\caption{Performance evaluation results on the ScanRefer, Nr3D, and Sr3D datasets.}
\label{tab:table1}
\scalebox{0.88}{
    \begin{tabular}{l|c|cccccccccccc}
    \toprule
    \multicolumn{1}{l|}{\multirow{3}{*}{Method}} & \multicolumn{1}{c|}{\multirow{3}{*}{Type}} & \multicolumn{6}{c}{ScanRefer}                                                                                                                                      & \multicolumn{3}{c}{Nr3D}                                                                                                             & \multicolumn{3}{c}{Sr3D}                                                                                                             \\ \cmidrule{3-14} 
    \multicolumn{1}{c|}{}                        & \multicolumn{1}{c|}{}                      & \multicolumn{2}{c}{Unique}                           & \multicolumn{2}{c}{Multiple}                         & \multicolumn{2}{c}{Overall}                          & \multicolumn{1}{c}{\multirow{2}{*}{Overall}} & \multicolumn{1}{c}{\multirow{2}{*}{Easy}} & \multicolumn{1}{c}{\multirow{2}{*}{Hard}} & \multicolumn{1}{c}{\multirow{2}{*}{Overall}} & \multicolumn{1}{c}{\multirow{2}{*}{Easy}} & \multicolumn{1}{c}{\multirow{2}{*}{Hard}} \\ \cmidrule{3-8}
    \multicolumn{1}{c|}{}                        & \multicolumn{1}{c|}{}                      & \multicolumn{1}{c}{@0.25} & \multicolumn{1}{c}{@0.5} & \multicolumn{1}{c}{@0.25} & \multicolumn{1}{c}{@0.5} & \multicolumn{1}{c}{@0.25} & \multicolumn{1}{c}{@0.5} & \multicolumn{1}{c}{}                         & \multicolumn{1}{c}{}                      & \multicolumn{1}{c}{}                      & \multicolumn{1}{c}{}                         & \multicolumn{1}{c}{}                      & \multicolumn{1}{c}{}                      \\ \midrule
ScanRefer \cite{chen2020scanrefer} & \multirow{15}{*}{Single} &
67.64 & 46.19 & 32.06 & 21.26 & 38.97 & 26.10    & 34.2 & 41.0 & 23.5 & -    & -    & -    \\
ReferIt3D \cite{achlioptas2020referit3d} & &
53.80 & 37.50 & 21.00 & 12.80 & 26.40 & 16.90    & 35.6 & 43.6 & 27.9 & 40.8 & 44.7 & 31.5 \\
TGNN \cite{huang2021text} & &
68.61 & 56.80 & 29.84 & 23.18 & 37.37 & 29.70    & 37.3 & 44.2 & 30.6 & 45.0 & 48.5 & 36.9 \\
InstanceRefer \cite{yuan2021instancerefer} & &
77.45 & 66.83 & 31.27 & 24.77 & 40.23 & 32.93    & 38.8 & 46.0 & 31.8 & 48.0 & 51.1 & 40.5 \\
SAT \cite{yang2021sat} & &
73.21 & 50.83 & 37.64 & 25.16 & 44.54 & 30.14    & 49.2 & 56.3 & 42.4 & 57.9 & 61.2 & 50.0 \\
FFL-3DOG \cite{feng2021free} & &
78.80 & 67.94 & 35.19 & 25.70 & 41.33 & 34.01    & 41.7 & 48.2 & 35.0 & -    & -    & -    \\
3DVG-Transformer \cite{zhao20213dvg} & &
77.16 & 58.47 & 38.38 & 28.70 & 45.90 & 34.47    & 40.8 & 48.5 & 34.8 & 51.4 & 54.2 & 44.9 \\
TransRefer3D \cite{he2021transrefer3d} & &
-     & -     & -     & -     & -     & -        & 42.1 & 48.5 & 36.0 & 57.4 & 60.5 & 50.2 \\
LanguageRefer \cite{roh2022languagerefer} & &
-     & -     & -     & -     & -     & -        & 43.9 & 51.0 & 36.6 & 56.0 & 58.9 & 49.3 \\
LAR \cite{bakr2022look} & &
-     & -     & -     & -     & -     & -        & 48.9 & 58.4 & 42.3 & 59.4 & 63.0 & 51.2 \\
3DJCG \cite{cai20223djcg} & &
78.75 & 61.30 & 40.13 & 30.08 & 47.62 & 36.14    & -    & -    & -    & -    & -    & -    \\
3D-SPS \cite{luo20223d} & &
81.63 & 64.77 & 39.48 & 29.61 & 47.65 & 36.43    & 51.5 & 58.1 & 45.1 & 62.6 & 56.2 & 65.4 \\
MVT \cite{huang2022multi} & &
77.67 & 66.45 & 31.92 & 25.26 & 40.80 & 33.26    & 55.1 & 61.3 & 49.1 & 64.5 & 66.9 & 58.8 \\
ViL3DRel \cite{chen2022language} & &
81.58 & 68.62 & 40.30 & 30.71 & 47.94 & 37.73    & 64.4 & 70.2 & 57.4 & 72.8 & 74.9 & 67.9 \\
BUTD-DETR $\dagger$ \cite{jain2022bottom} & &
81.93 & 67.11 & 42.61 & 31.84 & 50.24 & 38.68    & -    & -    & -    & -    & -    & -    \\
\hline
ViL3DRel + BS & \multirow{5}{*}{Dense} &
82.51 & 69.50 & 42.94 & 31.95 & 50.26 & 38.90    & 66.0 & 71.6 & 59.3 & 74.1 & 76.4 & 68.7 \\
BUTD-DETR + BS & &
84.63 & 68.90 & 45.41 & 33.68 & 53.02 & 40.51    & -    & -    & -    & -    & -    & -    \\
ViL3DRel + DepNet & &
87.34 & 71.04 & 47.56 & 35.93 & 54.92 & 42.43    & 70.6 & 75.8 & 64.4 & 78.0 & 80.5 & 72.2 \\
BUTD-DETR + DepNet & &
87.75 & 70.66 & 50.70 & 37.81 & 57.89 & 44.18    & -    & -    & -    & -    & -    & -    \\
3DOGSFormer (Ours) & &
\textbf{90.11} & \textbf{73.08} & \textbf{58.62} & \textbf{43.57} & \textbf{64.73} & \textbf{49.29}    & \textbf{73.5} & \textbf{77.9} & \textbf{68.2} & \textbf{80.8} & \textbf{83.4} & \textbf{74.7} \\
\bottomrule
\end{tabular}
}
\end{table*}

With the existing 3D single-object grounding datasets, we develop a strategy to simulate the proposed 3D dense object grounding setting where humans are likely to describe multiple nearby objects in a focused region. Specifically, for each training step, we first randomly sample a 3D scene from the training set. Next, we select a random object in the scene as the focused object and randomly choose its $K-1$ nearest objects as the concerned objects. 
At last, we sample a referring sentence for each of the $K$ ($2\leq K \leq 12$) selected objects, and then we arrange these sentences into a paragraph in such a way that referring sentences for objects closer to the center object have higher probability to appear earlier in the paragraph.
Paragraphs with fewer than 12 sentences are padded with zeros to 12 sentences. Such a random sampling strategy reduces the model to overfit spatial relation priors of dense objects. During the evaluation, we divide sentences in the dataset into paragraphs in the same way so that the target objects of the sentences in the same paragraph form a KNN cluster in the 3D scene. We mainly sample 12 sentences per paragraph for evaluation.

\noindent\textbf{Comparison Baselines.}
We compare the proposed 3DOGSFormer model with most of the existing 3D single-object grounding methods since we can adapt these methods to the dense grounding setting straightforwardly by applying them to each sentence in the paragraph. To better understand our 3DOGSFormer's performance, we further extend two of the state-of-the-art 3D single-object grounding models, \textbf{ViL3DRel} \cite{chen2022language} and \textbf{BUTD-DETR} \cite{jain2022bottom}, to the 3D DOG setting as more competitive baselines. We devise two strategies to extend these models by leveraging \textbf{Beam Search (BS)} and \textbf{DepNet} \cite{bao2021dense} (a dense grounding method in the 2D field), respectively. In specific, the BS model first localizes each sentence in the 3D scene independently with the base model, then applies beam search on the top 12 grounding results of each object as post-processing, such that the final dense object grounding results are most concentrated in the 3D scene, measured by center variances. In the DepNet-based strategy, we first extract features before the grounding head of the base model for each referring sentence, and then apply DepNet's DEAP module \cite{bao2021dense} to produce enhanced features via global reasoning, which are fed into the grounding head to generate grounding results. We adopt a pretrain-then-finetune strategy to optimize the extended model in this case. At last, we obtain 4 combined strong baselines: \textbf{ViL3DRel+BS}, \textbf{BUTD-DETR+BS}, \textbf{ViL3DRel+DepNet}, and \textbf{BUTD-DETR+DepNet}.

\subsection{Comparison Results}

Table \ref{tab:table1} shows the comparison results for all methods on ScanRefer, Nr3D, and Sr3D. $\dagger$ denotes that we reevaluate BUTD-DETR in a fair setting since its reported performance ignores some challenging samples. We only evaluate BUTD-DETR and its variants on ScanRefer because it needs to regress the target bounding box while models in the Nr3D/Sr3D setting identify the target among GT boxes. We modify our 3DOGSFormer accordingly to evaluate on Nr3D/Sr3D by replacing the 3D scene encoder with a GT box encoder and modifying the decoders to select the target GT boxes. The comparison results reveal some interesting points. (1) All 3D DOG methods outperform 3D single-object grounding methods with a clear margin, and even a simple BS strategy can improve the SOTA single-object grounding models. It is due to the fact that we can access more information about the densely referred objects in the 3D DOG setting to design our systems for more accurate grounding, which verifies the superiority of the proposed 3D DOG setting. (2) The $\{\cdot\}$+DepNet methods achieve clearly superior results than their base models and the $\{\cdot\}$+BS methods. This is because the additional global reasoning module enables the base model to capture the paragraph-level information while the SOTA single grounding methods can only conduct contextual modeling of a single target object. The performance gap validates the necessity of jointly modeling multiple target objects in a paragraph in 3D DOG. (3) Our 3DOGSFormer outperforms all baselines, especially the competitive $\{\cdot\}$+DepNet methods, with a significant margin, which suggests that the proposed 3DOGSFormer framework can efficiently capture the semantic relations among multiple sentences by the contextual query generator and our proposal-guided global transformer decoder can predict the target boxes precisely via modeling the 3D spatial relations among densely referred objects.

\begin{table}[!t]
\center
\caption{Ablation study of the 3DOGSFormer components.}
\label{tab:table2}
\scalebox{0.88}{
\begin{tabular}{lccc|ccc}
\toprule
\makebox[0.1\linewidth][c]{\ } & \makebox[0.1\linewidth][c]{CQG} & \makebox[0.1\linewidth][c]{LTD} & \makebox[0.1\linewidth][c]{GTD} & \multicolumn{1}{c}{\begin{tabular}[c]{@{}c@{}}Unique\\ @0.5\end{tabular}} & \multicolumn{1}{c}{\begin{tabular}[c]{@{}c@{}}Multiple\\ @0.5\end{tabular}} & \multicolumn{1}{c}{\begin{tabular}[c]{@{}c@{}}Overall\\ @0.5\end{tabular}} \\
\midrule
R1 &            & \checkmark &            & 66.51 & 31.67 & 38.43 \\
R2 & \checkmark & \checkmark &            & 68.92 & 36.59 & 42.86 \\
R3 & \checkmark &            & \checkmark & 70.88 & 39.53 & 45.61 \\
R4 &            & \checkmark & \checkmark & 71.85 & 41.64 & 47.50 \\
R5 & \checkmark & \checkmark & \checkmark & \textbf{73.08} & \textbf{43.57} & \textbf{49.29} \\ 
\bottomrule
\end{tabular}
}
\end{table}

\begin{table}[!t]
\center
\caption{Ablation study of the Global Transformer Decoder.}
\label{tab:table3}
\scalebox{0.88}{
\begin{tabular}{lccc|ccc}
\toprule
\makebox[0.1\linewidth][c]{\ } & \makebox[0.1\linewidth][c]{$A^E$} & \makebox[0.1\linewidth][c]{$A^I$} & \makebox[0.1\linewidth][c]{$A^F$} & \multicolumn{1}{c}{\begin{tabular}[c]{@{}c@{}}Unique\\ @0.5\end{tabular}} & \multicolumn{1}{c}{\begin{tabular}[c]{@{}c@{}}Multiple\\ @0.5\end{tabular}} & \multicolumn{1}{c}{\begin{tabular}[c]{@{}c@{}}Overall\\ @0.5\end{tabular}} \\
\midrule
R1&            &        &      & 70.97 & 39.80 & 45.85 \\
R2& \checkmark &        &      & 71.81 & 41.13 & 47.08 \\
R3&          & \checkmark &      & 71.29 & 40.20 & 46.23 \\
R4&          &      & \checkmark & 71.67 & 40.52 & 46.56 \\
R5& \checkmark & \checkmark &      & 72.01 & 42.21 & 47.99 \\
R6&          & \checkmark & \checkmark & 72.15 & 42.34 & 48.12 \\
R7& \checkmark &      & \checkmark & 72.65 & 42.94 & 48.70 \\
R8& \checkmark & \checkmark & \checkmark & \textbf{73.08} & \textbf{43.57} & \textbf{49.29} \\ 
\bottomrule
\end{tabular}
}
\end{table}

\subsection{Ablation Study}

To investigate the effectiveness of each component in 3DOGSFormer, we conduct ablation studies on ScanRefer. Concretely, our 3DOGSFormer includes the Contextual Query Generator (CQG), Local Transformer Decoder (LTD), and Global Transformer Decoder (GTD). We selectively discard them to generate ablation models and report the results in Table \ref{tab:table2}. From these results, we can find that the full model outperforms all ablation models, validating each component is helpful for 3D dense object grounding. The baseline R1 using only LTD achieves comparable results as BUTD-DETR since they have similar DETR-based architectures. Rows R2-R4 show that removing GTD causes the worst performance degradation, demonstrating the 3D spatial relation understanding captured by GTD is the most essential in 3D DOG. Comparing R2 to R1 and R5 to R4, we find that adding CQG can consistently improve the performance, showing the effectiveness of modeling semantic relations of multiple sentences.

Since GTD is critical in 3DOGSFormer, we further conduct detailed ablation studies on the key modules in GTD, the proposal-guided attention modules, consisting of the explicit spatial matrix $A^E$, implicit spatial matrix $A^I$, and focused region constraint matrix $A^F$. We discard them selectively to create ablation models. As shown in Table \ref{tab:table3}, the full model achieves the best performance, verifying all the proximity and constraint matrices are effective in 3D DOG. If only one matrix is applied, the model with $A^E$ is the best, demonstrating the explicit spatial modeling is the most important to capture 3D spatial relations among dense objects. And if two matrices are used, the model with $A^E$ and $A^F$ outperforms other models, suggesting the focused region modeling is crucial in 3D spatial relation understanding and high-quality dense grounding.

\begin{figure}[!t]
  \setlength\tabcolsep{0.1 pt}
  {\renewcommand{\arraystretch}{1}
    \begin{tabular}{ccc}
      \centering
      \includegraphics[width=0.23\textwidth]{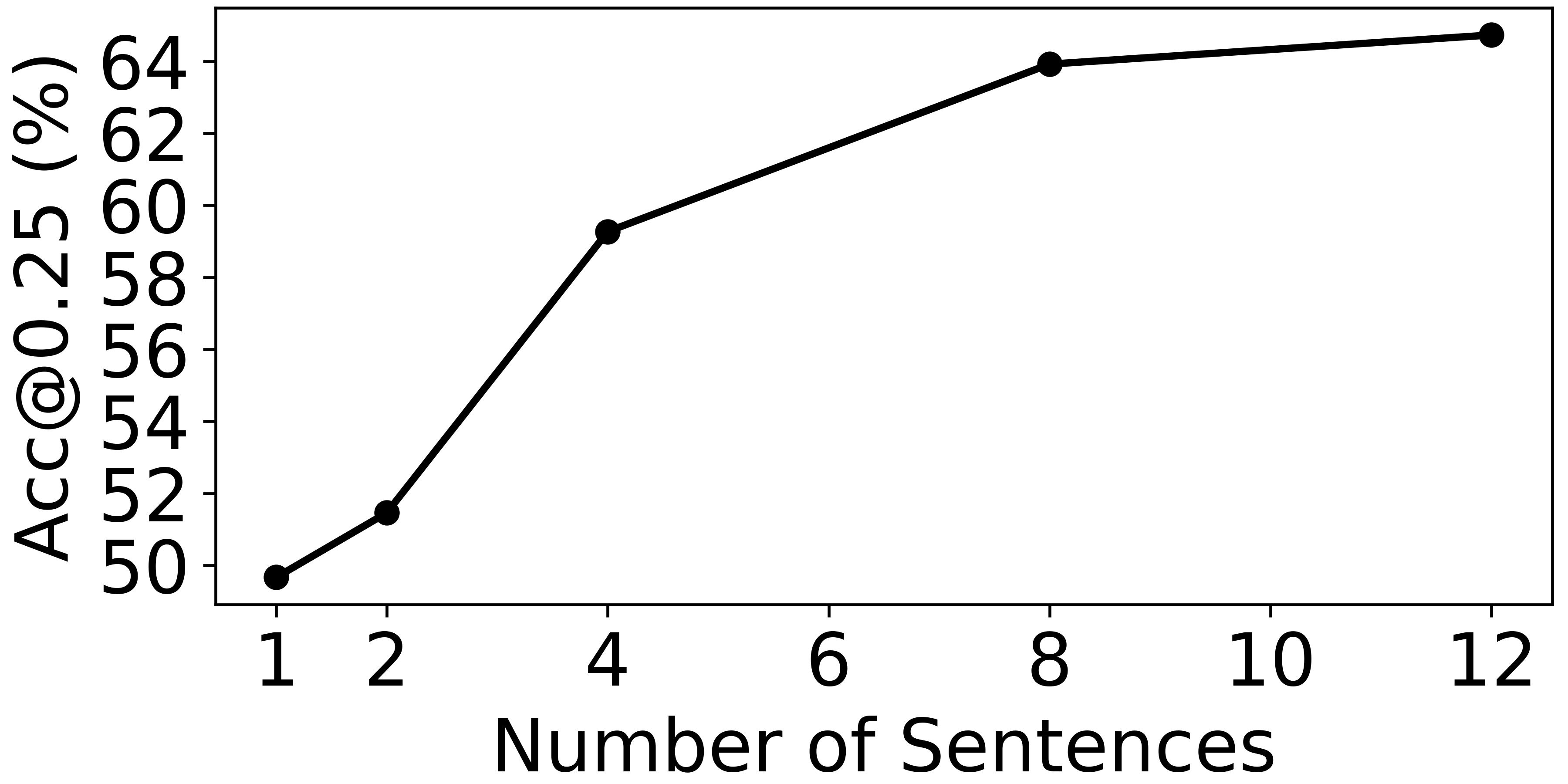} & \ \ & \includegraphics[width=0.23\textwidth]{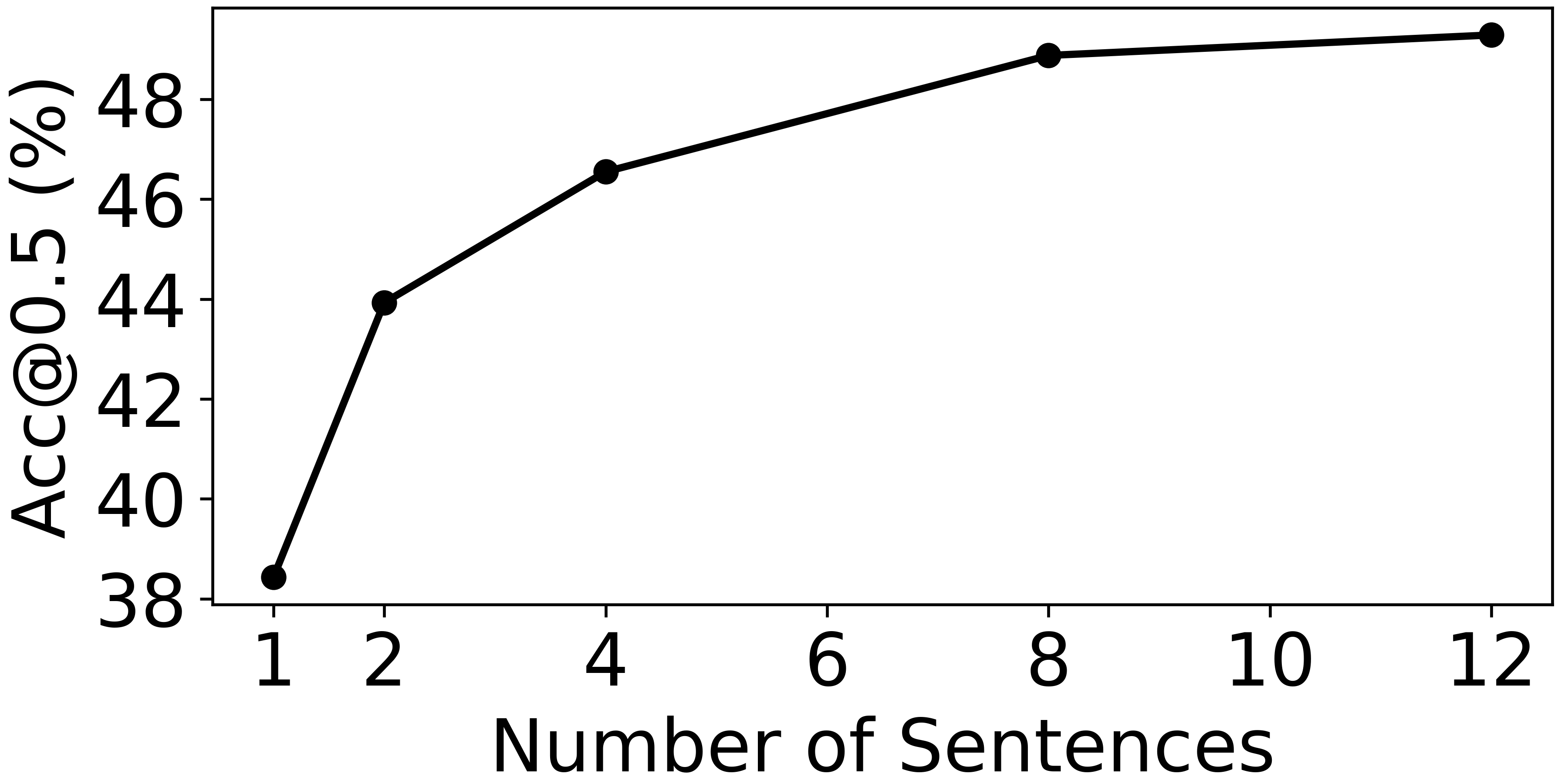}\\
    \end{tabular}
    \vspace{-0.1cm}
  }
  \smallskip
  \caption{Effect of the number of sentences in paragraphs.}
  \label{fig:ablation} 
\end{figure}

Moreover, we explore the effect of the number of sentences in paragraphs as shown in Figure \ref{fig:ablation}. Specifically, we evaluate the well-trained 3DOGSFormer on reconstructed ScanRefer validation sets by sampling with different numbers of sentences in paragraph descriptions. We report the overall performances. As can be seen, the metrics achieve higher values when the number of sentences increases, validating that descriptions with more sentences can provide richer semantic and spatial information for our model to obtain more precise grounding.

\subsection{Qualitative Results}
\begin{figure}[!tbp]
    \begin{center}
    \includegraphics[width=0.475\textwidth]{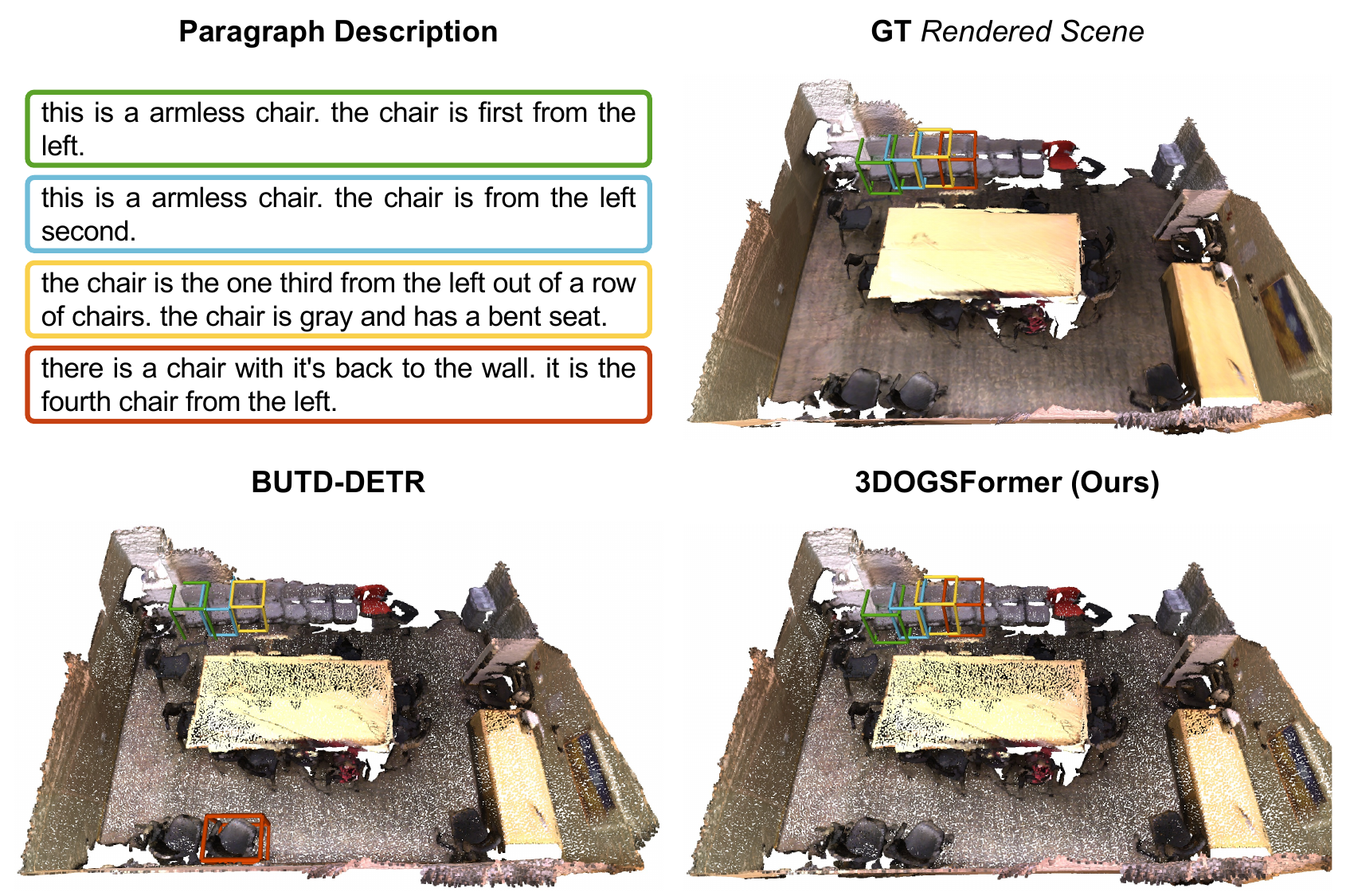}
    \end{center}
    \caption{An example of 3D Dense Object Grounding results.}
    \label{fig:demo}
\end{figure}

We qualitatively compare 3DOGSFormer to the baseline model BUTD-DETR on ScanRefer and display a typical example in Figure \ref{fig:demo}. By capturing additional semantic and spatial relation information of dense target objects, 3DOGSFormer produces more precise results.

\section{Conclusion}
In this paper, we propose a novel 3D DOG task to explore the 3D dense object grounding. We devise a 3DOGSFormer model to tackle 3D DOG in a two-phase grounding pipeline. In the first phase, it initializes the grounding proposals with a local transformer decoder, where a contextual query generator is developed to efficiently capture the semantic relationships among the paragraph. In the second phase, it refines the initial proposals with a global transformer decoder, which contains newly designed proposal-guided attention layers to improve 3D spatial relation understanding via encoding pairwise spatial relations both explicitly and implicitly. Extensive experiments on three benchmarks demonstrate the significance of our 3DOGSFormer framework.


\bibliographystyle{ACM-Reference-Format}
\balance
\bibliography{main}

\appendix

\end{document}